\documentclass{article}

\usepackage{arxiv}

\usepackage{amsmath,amsfonts,bm}

\def\eqref#1{equation~\ref{#1}}

\def\1{\bm{1}}

\DeclareMathAlphabet{\mathsfit}{\encodingdefault}{\sfdefault}{m}{sl}
\SetMathAlphabet{\mathsfit}{bold}{\encodingdefault}{\sfdefault}{bx}{n}

\usepackage{hyperref}
\usepackage{url}
\usepackage{amsmath}
\usepackage{amssymb}
\usepackage[pdftex]{graphicx}
\usepackage{amsmath}
\usepackage[utf8]{inputenc} 
\usepackage[T1]{fontenc}    
\usepackage{hyperref}       
\usepackage{booktabs}       
\usepackage{amsfonts}       
\usepackage{nicefrac}       
\usepackage{microtype}      
\usepackage{lipsum}
\usepackage{graphicx}
\usepackage{soul,color}
\usepackage{appendix}
\usepackage{multirow}
\usepackage{threeparttable}

\newcommand{\TheName}{\textbf{\texttt{SMILE}}}
\usepackage{algorithm}
\usepackage{algpseudocode}
\usepackage{stfloats}

\title{SMILE: \underline{S}elf-Distilled \underline{MI}xup for Efficient Transfer \underline{LE}arning}

\author{Xingjian Li$^{1,2}$\footnotemark[1], Haoyi Xiong$^{1}$\footnotemark[1], Cheng-Zhong Xu$^{2}$, Dejing Dou$^{1}$\\
\\
{\small $^1$Big Data Lab, Baidu Research,} \\{\small $^2$State Key Lab of IOTSC, Department of Computer Science, University of Macau}\\
{\small \{lixingjian,xionghaoyi,doudejing\}@baidu.com, czxu@um.edu.mo}}

\begin{document}
\renewcommand{\thefootnote}{\fnsymbol{footnote}} \footnotetext[1]{Equal contribution.}

\maketitle
\begin{abstract}
To improve the performance of deep learning, mixup has been proposed to force the neural networks favoring simple linear behaviors in-between training samples. Performing mixup for transfer learning with pre-trained models however is not that simple,  a high capacity pre-trained model with a large fully-connected (FC) layer could easily overfit to the target dataset even with samples-to-labels mixed up. In this work, we propose \TheName — \underline{S}elf-Distilled \underline{M}ixup for Eff\underline{I}cient Transfer \underline{LE}arning. With mixed images as inputs, \TheName\ regularizes the outputs of CNN feature extractors to learn from the mixed feature vectors of inputs (sample-to-feature mixup), in addition to the mixed labels. Specifically, \TheName\ incorporates a mean teacher, inherited from the pre-trained model, to provide the feature vectors of input samples in a self-distilling fashion, and mixes up the feature vectors accordingly via a novel triplet regularizer. The triple regularizer balances the mixup effects in both feature and label spaces while bounding the linearity in-between samples for pre-training tasks. Extensive experiments have been done to verify the performance improvement made by \TheName, in comparisons with a wide spectrum of transfer learning algorithms, including fine-tuning, L2-SP, DELTA, and RIFLE, even with mixup strategies combined. Ablation studies show that the vanilla sample-to-label mixup strategies could marginally increase the linearity in-between training samples but lack of generalizability, while \TheName\ significantly improve the mixup effects in both label and feature spaces with both training and testing datasets. The empirical observations backup our design intuition and purposes. 
\end{abstract}

\section{Introduction}
Performance of deep learning algorithms in real-world applications is often limited by the size of training datasets. Training a deep neural network (DNN) model with a small number of training samples usually leads to the \emph{over-fitting} issue with poor generalization performance. A common yet effective solution is to train DNN models under transfer learning~\cite{pan2010survey} settings using large source datasets. The knowledge transfer from the source domain helps DNNs learn better features and acquire higher generalization performance for the pattern recognition in the target domain~\cite{donahue2014decaf,yim2017gift}.

In addition to deep transfer learning, another effective strategy for improve generalization performance of DNN is mixup~\cite{zhang2018mixup}, where the objective is to have DNNs in the learning procedure favor the \emph{linear behaviors} in-between training samples. To achieve the goal, the mixup strategy picks up multiple images from the training set, mixes the samples and labels proportionally to generate a new pair of sample and label for data augmentation. The regularization effects brought by \TheName\ could help control the complexity of DNN models~\cite{hanin2019complexity,vapnik2013nature} while largely improving the robustness and generalization performance~\cite{zhang2020does}. However, considering the strong capacity of pre-trained models and the limited training dataset, our research yields the concern as follow,
\begin{itemize}\vspace{-3mm}
    \item[] \emph{Can fine-tuning with pre-trained models overfit to the mixed-up samples and labels?}\vspace{-3mm}
\end{itemize}

\textbf{Our Observations. } We find fine-tuning with high capacity pre-trained models CAN overfit to the mixup samples/labels. From mixup, we simply derive a linear interpolation loss to measure the error of linear interpolation between a pair of samples $(x_1,y_1)$ and $(x_2,y_2)$ for the model $f(\cdot)$,
\begin{equation}
    \left\|f(\lambda x_1+(1-\lambda) x_2)-(\lambda f(x_1)+(1-\lambda) f(x_2))\right\|_2^2\ ,
\end{equation}
where a lower linear interpolation loss indicates stronger linear behaviors in-between the samples and usually better generalization performance~\cite{zhang2020does}. 
Our experiments however find that fine-tuning with mixup could obtain a low interpolation loss in the training dataset while suffering a high interpolation loss in the test set ($\geq$25\% higher interpolation loss on the testing set than the one on training set, please see also in \textbf{Section 5}). Compared to fine-tuning without mixup on the target domain, fine-tuning with mixup improves the performance of DNNs with reduced margins. This observation indicates that the \emph{linear behaviors} gained by mixup regularization could not well generalize to the testing dataset and overfit to the mixup samples/labels from the training dataset. Thus, our research intends to study a way to \emph{make mixup strategies generalizable in deep transfer learning settings while significantly improving the performance of DNNs.}

To achieve the above goal, some non-trivial technical challenges should be tackled. 

\underline{\emph{(I) Sample-to-Feature Mixup}}. A high-capacity pre-trained model, offering tons of well-trained features, would force a Fully-Connected (FC) Layer to memorize samples and labels mixed-up with trivial updates to weights of its CNN feature extractor. Though some randomized strategies, such as RIFLE~\cite{li2020rifle}, could deepen back-propagation in vanilla transfer learning settings, it is still challenging to reinforce the mixup effects in the CNN feature extractor. 
    
\underline{\emph{(II) Mixed-up Feature Vectors}}. To ensure mixup effects in outputs of CNN feature extractors, a possible way is to let CNN feature extractors learn from the mixed-up samples and feature vectors, while the ground-truth feature vectors are usually not available. Thus, there needs to accurately estimate the feature vector for any sample in the target dataset before having the CNN trained (i.e., a \emph{Chicken or the Egg} problem).
    
\underline{\emph{(III) Cross-Domain Generalizability}}. A pre-trained DNN usually is capable of behaving linearly under interpolation of the source dataset. During the fine-tuning procedure, it is reasonable to doubt that such linear behaviors in source domain might be forgotten~\cite{chen2019catastrophic}. To improve the generalization performance, there thus needs to preserve the linear behaviors in the source domain and transfer such ability to the target domain during fine-tuning.
    

\textbf{Our Work.} To address above technical challenges, in this work, we propose \TheName---Self-distilled Mixup strategies for Efficient Transfer Learning. Instead of regularizing mixup effects in label spaces, \TheName\ regularizes the outputs of CNN feature extractor with mixed-up samples and feature vectors, where the feature vectors are extracted from a \emph{mean teacher}. Specifically, the \emph{mean teacher} is initialized by the pre-trained model and is updated from the training CNN (the \emph{student model}) after every $10$ iterations of the fine-tuning process. Finally, to ensure the performance of mixup, \TheName\ proposes the triplet loss for regularization: \textbf{(1)} the Euclidean distance between mixed-up feature vectors (extracted from the \emph{mean teacher}) and the feature vector of mixed-up samples (the CNN output of the \emph{student model}) has been used to ensure the feature-wise mixup; \textbf{(2)} in addition to the feature space, the vanilla sample-to-label mixup loss on the target dataset has been used to promote the linear behaviors in the \emph{student model}; and \textbf{(3)} both the \emph{mean teacher} and \emph{student models} train an individual FC-layer classifier to adapt the original source dataset, while the \emph{student model} uses vanilla sample-to-label mixup loss on the source dataset as a regularizer to make the linear behaviors in source domain preserved.

%
To the best of our knowledge, this work has made three sets of contributions as follows.

\textbf{\em (1)}~ We study the problem of regularizing DNNs to enjoy mixup effects under deep transfer learning settings, where the major concerns is to avoid the overfitting to mixed-up samples and labels, using a high-capacity pre-trained model but with a small target training dataset. We elaborate the technical issues, and propose to solve the problem through enabling sample-to-feature mixup, where obtaining the feature vectors for mixup and ensuring cross-domain generalizability of linear behaviors become the key challenges.

\textbf{\em (2)} We propose \TheName---\underline{s}elf-distilled \underline{mi}xup for efficient transfer \underline{le}arning, where self-distillation with teacher-student networks has been used as the core framework of the solution. Specifically, \TheName\ leverages triplet loss to regularize the student network. Given two samples drawn from the target domain as the input, \TheName\ linearly combines two samples proportionally and sends the mixed-up sample to the student network. It constrains the Euclidean distance between the output of teacher model's CNN feature extractor and  a mixed-up feature vector (i.e., linear combination of the teacher model's outputs for the two samples) via a \emph{sample-to-feature mixup}. Then, \TheName\ regularizes classification results of the student model via vanilla sample-to-label mixup using the mixed-up label (i.e., linear combination of ground-truth labels). Further, to obtain cross-domain generalizability, \TheName\ trains an additional FC classifier for both teacher and student networks to adapt the target dataset but in the source domain. It regularizes student network using self-distilled sample-to-label mixup to learn from the linear combination of teacher model's classification results on source domain.

\textbf{\em (3)} We carry out extensive experiments using a wide range of source and target datasets, and compare the results of \TheName\ with a number of baseline algorithms, including fine-tuning with weight decay ($L^2$)~\cite{donahue2014decaf}, fine-tuning with $L^2$-regularization on the starting point ($L^{2}$-$SP$)~\cite{li2018explicit}, DELTA~\cite{li2019delta}, Batch Singular Shrinkage (BSS)~\cite{chen2019catastrophic}, RIFLE~\cite{li2020rifle},  and Co-Tuning~\cite{you2020co} with/without mixup strategies. The experiment results showed that \TheName\ can outperform all these algorithms with significant improvement in both efficiency and effectiveness. The ablation studies show that (1) sample-to-feature mixup design is significantly better than vanilla sample-to-label mixup for deep transfer learning; (2) performing both sample-to-feature and sample-to-label mixup on the target training dataset is much better than perform one of these two; and (3) the proposed self-distilled sample-to-label mixup on the source domain can further improve the generalization performance.

    

\textbf{Organizations of the Paper} The rest of this paper is organized as follows. In Section~2, we review related work, where the most relevant studies are discussed. We present the algorithm design in Section~3, and the experiments with overall comparison results in Section~4, respectively. We discuss the details about the algorithm with case studies and ablation studies in Section~5. We conclude the paper in Section 6. 

\section{Related Work}
In this section, we first introduce the related works from deep transfer learning's perspectives, then we discuss the most relevant work to our study.

\subsection{Deep Transfer Learning}
To enable transfer learning for DNNs, finetuning~\cite{donahue2014decaf} has been proposed to first train a DNN model using the large (and  possibly irrelevant) source dataset (e.g. ImageNet), then uses the weights of the pre-trained model as the starting point of optimization and fine-tunes the model using the target dataset. In this way, blessed by the power of large source datasets, the fine-tuned model is usually capable of handling the target task with better generalization performance. Furthermore, authors in~\cite{yim2017gift,li2018explicit,li2019delta} propose transfer learning algorithms that regularize the training procedure using the pre-trained models, so as to constrain the divergence of the weights and feature maps between the pre-trained and fine-tuned DNN models. Later, the work~\cite{chen2019catastrophic,wan2019towards} introduces new algorithms that prevent the regularization from the hurts to transfer learning, where ~\cite{chen2019catastrophic} proposes to truncate the tail spectrum of the batch of gradients while~\cite{wan2019towards} proposes to truncate the ill-posed direction of the aggregated gradients. In addition to the aforementioned strategies, multi-tasking algorithms have been used for deep transfer learning, such as~\cite{ge2017cvpr,cui2018large}. 


While all above algorithms enable knowledge transfer from source datasets to target tasks, they unfortunately perform poorly due to the catastrophic forgetting and negative transfer. Most transfer learning algorithms~\cite{donahue2014decaf,yim2017gift,li2018explicit,li2019delta} consist of two steps -- pre-training and fine-tuning. Given the features that have been learned in the pre-trained models, either forgetting some good features during the fine-tuning process (\emph{catastrophic forgetting})~\cite{chen2019catastrophic} or preserving the inappropriate features/filters to reject the knowledge from the target domain (\emph{negative transfer})~\cite{li2019delta,wan2019towards} would hurt the performance of transfer learning. In this way, there might need a way to make compromises between the features learned from both source/target domains during the fine-tuning process, where multi-task learning with Seq-Train~\cite{cui2018large} and Co-Train~\cite{ge2017cvpr} might suggest feasible solutions to well-balance the knowledge learned from the source/target domains, through fine-tuning the model with a selected set of auxiliary samples (rather than the whole source dataset)~\cite{cui2018large} or alternatively learning the features from both domains during fine-tuning~\cite{ge2017cvpr}.

\subsection{Connections to Our work}
The most relevant studies to our algorithm are~\cite{verma2019manifold,yun2019cutmix,li2019delta,chen2019catastrophic,li2020rifle}. While the first two works~\cite{verma2019manifold,yun2019cutmix} propose to improve mixup and its derivatives for data augmentation through interpolating the feature spaces, the rest three works~\cite{li2019delta,chen2019catastrophic,li2020rifle} focus on improving deep transfer learning through regularizing the feature spaces.  


The manifold mixup strategy~\cite{verma2019manifold} has been proposed to smooth the decision boundary of DNN classifiers using mixed-up feature maps and labels, in a \emph{feature-to-label} mixup fashion. 
On the other hand, MixCut~\cite{yun2019cutmix} also proposes a \emph{sample-to-feature} data augmentation strategy, where the algorithm fuses two images into one and forms a new feature map accordingly, with respect to the localizable visual features in two images. Compared to above works, the major technical difficulty of \TheName\ is that above algorithms use feature maps extracted from CNN models directly, while \TheName\ regularizes the output of CNN feature extractor when accurate estimates of feature vectors are not available (the CNN is under fine-tuning to adapt the target dataset).


While~\cite{li2019delta,chen2019catastrophic} propose to improve the feature-wise knowledge distillation or spectral regularization for transfer learning,~\cite{li2020rifle} studies way to regularize the pre-trained CNN feature extractor, during fine-tuning, by incorporating randomness from FC layers. Compared to above algorithms, \TheName\ is proposed to solve the problem of overfitting to mixup under deep transfer learning settings. Our ablation studies in Section 5 show that the simple combination of fine-tuning and mixup strategies does not well in transfer learning settings from both linear behaviors preservation and generalization performance aspects. \TheName\ makes unique contributions in proposing novel \emph{self-distilled sample-to-feature mixup} strategies to improve performance in transfer learning with pre-trained models.

\begin{figure*}[t]
\centering
\includegraphics[width=0.95\textwidth]{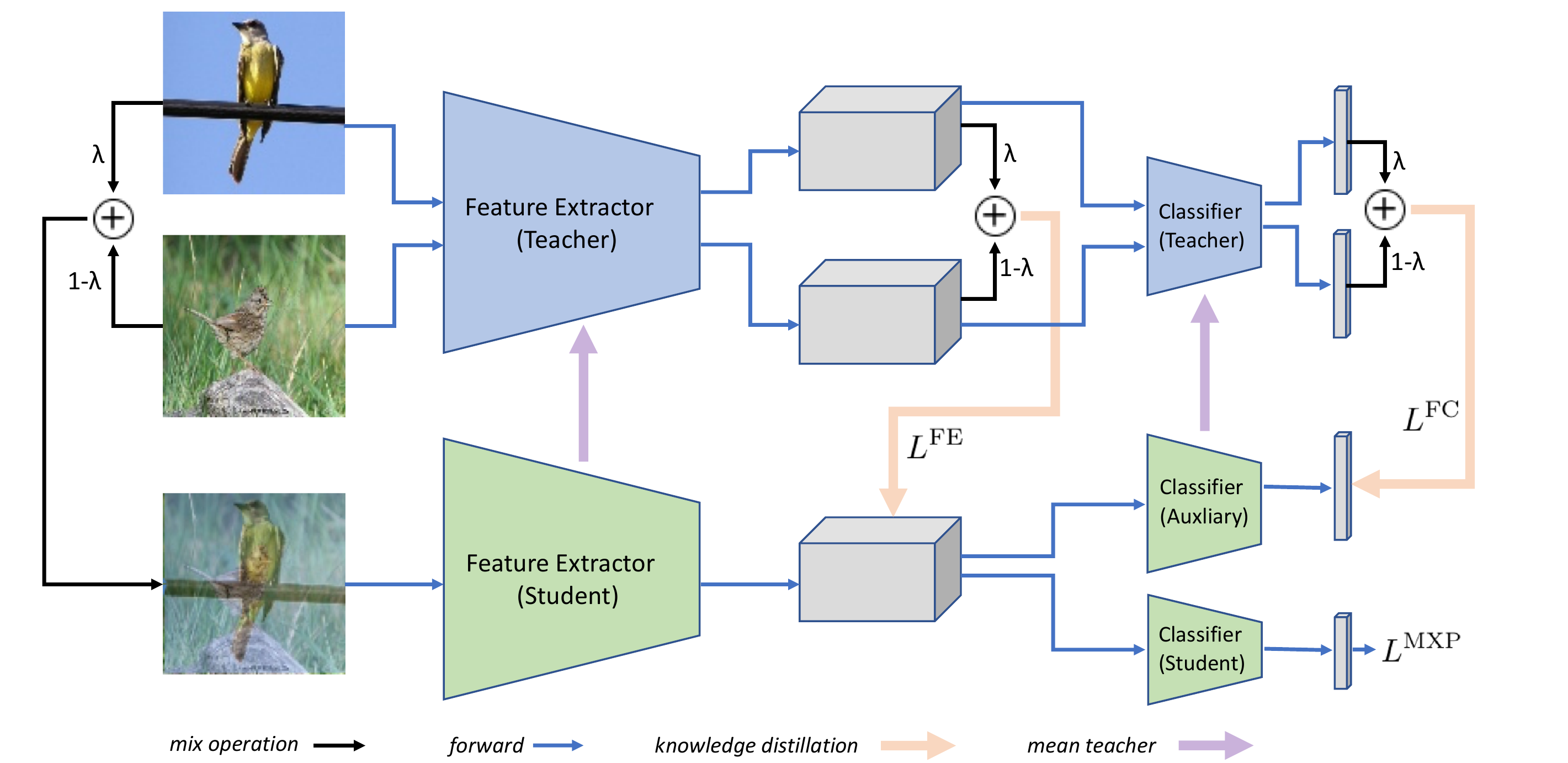} 
\caption{The Architecture of \TheName: Mean Teacher and Student Training with Triplet Regularization.}
\label{fig:arch} 
\vspace{-3mm}
\end{figure*}

\section{\TheName: Self-distilled Sample-to-Feature Mixup for Efficient Transfer Learning}
In this section, we first introduce the overall framework of \TheName, where the architectures of mean teacher-student training with triplet regularization is presented. Then, we specify the design of triplet regularizer and discuss the mixup effects incorporated by \TheName.

\begin{algorithm}
    \caption{Mean Teacher-Student Training of \TheName}
    \begin{algorithmic}[1]
    \Procedure{\TheName}{$\mathbf{D}^\mathrm{tgt},\mathbf{\omega}_\mathrm{src}, \eta, P, K$}
    \State $\omega_s^0\gets\mathbf{\omega}_\mathrm{src}$
     \For{$k = 1, 2, 3..., K$}
     \State \textcolor{blue}{/*Updating the Mean Teacher Model*/}
     \State $\kappa\gets k ~\mathbf{mod}~ P$
     \If{$\kappa = 0$}
     \State $\mathbf{\omega}_t\gets\omega^{k-1}$

     \EndIf
     \State \textcolor{blue}{/*Updating the Student Model with SGD*/}
     \State $\mathbf{B}_t\gets\textbf{ mini-batch sampling from}~\mathbf{D}^\mathrm{tgt}$
     \State $\mathbf{g}^k\gets\frac{1}{|\mathbf{B}_t|}\sum_{(x,y)\in\mathbf{B}_t}\nabla\mathcal{L}_{(x,y)}(\omega_s^k,\mathbf{\omega}_t)$
     \State $\omega_s^{k}\gets\omega_s^{k-1}-\eta\cdot \mathbf{g}^k$
     \EndFor
     \State \Return{$\omega^K$}
     \EndProcedure
    \end{algorithmic}
    \label{alg:smilelearn}
\end{algorithm}

\subsection{Overall Framework}
Given a target training dataset  $\mathbf{D}^{\mathrm{tgt}}=\{({x}_1,y_1),({x}_2,y_2)... ({x}_n,y_n)\}$ and a model $\mathbf{\omega}_\mathrm{src}$ pre-trained with the source dataset, \TheName\ learns a model $\mathbf{\omega}$ to adapt the target dataset in a mean teacher-student training procedure. 

Algorithm~\ref{alg:smilelearn} presents the design of the mean teacher-student training procedure. Specifically, \TheName\ initializes both teacher $\omega_t$ and student $\omega_s$ model with the pre-trained model $\omega_\mathrm{src}$ and updates the mean teacher model from the student model for every $P$ iterations ($P$ is set to $10$ iterations in our research).
Furthermore, with the teacher and student models (i.e., $\omega_t$ and $\omega^k_s$ in the $k^{th}$ iteration), \TheName\ updates the student model through minimizing a loss function as follow,
\begin{equation} \label{eq:smile}
\underset{w}{\mathrm{min}}\ \left\{\mathcal{L}(\omega,\omega_t)= \frac{1}{n}\sum_{i=1}^n L(z( {x}_{i}, \mathbf{\omega}), y_{i}) +  L^\mathrm{Tri}(\mathbf{\omega},\mathbf{\omega}_t)\right\}\ ,
\end{equation}
where 
$L(z( {x}_{i}, \mathbf{\omega}), y_{i})$ refers to the training loss of $(x_i,y_i)$ based on the model $\mathbf{\omega}$, and $L^\mathrm{Tri}(\mathbf{\omega},\mathbf{\omega}_t)$ refers to the loss for triplet regularization based on the teacher model $\mathbf{\omega}_t$. Note that the computation of triplet regularization $L^\mathrm{Tri}(\mathbf{\omega},\mathbf{\omega}_t)$ adopts the source dataset $\mathbf{D}^\mathrm{src}$ and two classifiers $z'_\mathrm{src}$ and $z''_\mathrm{src}$ as part of inputs and components. After $K$ iterations, \TheName\ outputs the student model as the overall result.

\subsection{Deep Transfer Learning with Regularization} 
Figure~\ref{fig:arch} presents the architecture of \TheName, where three losses for self-distillation between the mean teacher and student models form the triplet regularization as follow,
\begin{equation}\label{eq:triplet}
\begin{aligned}
    L^\mathrm{Tri}(\omega,\omega_t)=&\gamma_\mathrm{FE}\cdot L^\mathrm{FE}(\omega,\omega_t) + \gamma_\mathrm{FC}\cdot L^\mathrm{FC}(\omega,\omega_t)+L^\mathrm{MXP}(\omega),
\end{aligned}
\end{equation}
where $\gamma_\mathrm{FE}$ and $\gamma_\mathrm{FC}$ refer to the weight of the two terms, the term $L^\mathrm{FE}(\omega,\omega_t)$ refers to the sample-to-feature mixup regularizer based on student and teacher models on target domain $\mathbf{D}^\mathrm{tgt}$, the term $L^\mathrm{FC}(\omega,\omega_t)$ refers to the sample-to-label mixup regularizer based on student and teacher models on the source domain (e.g., 1000 classes when the model was pre-trained using ImageNet), and the term $L^\mathrm{MXP}(\omega)$ refers to the vanilla sample-to-label mixup regularizer based on student model on the target domain $\mathbf{D}^\mathrm{tgt}$. 

Specifically, the sample-to-feature mixup regularizer based on teacher and student models on the target domain defined as
\begin{equation}
    \begin{aligned}
        L^\mathrm{FE}(\mathbf{\omega},\mathbf{\omega}_t)
        =\underset{\lambda\sim\mathrm{Beta}(\alpha,\alpha)}{\mathbb{E}}\ \underset{x_i,x_j\sim\mathbf{D}^\mathrm{tgt}}{\mathbb{E}} ||\mathrm{FE}\left(\mathrm{Mix}_\lambda(x_i, x_j);\mathbf{\omega}\right)
        - \mathrm{Mix}_\lambda(\mathrm{FE}(x_i;\mathbf{\omega}_{t}),\mathrm{FE}(x_j;\mathbf{\omega}_{t}))||_2^2\ ,\label{eq:kd}
    \end{aligned}
    \end{equation}
where the operator $\mathrm{Mix}_\lambda(u,v)=(1-\lambda)\cdot u+\lambda\cdot v$ referring to the linear combination of two inputs, $\lambda\sim\mathrm{Beta}(\alpha,\alpha)$ proposed by~\cite{zhang2018mixup}  refers to the linear combination coefficient drawn from a Beta distribution, and $\mathrm{FM}(x_i;\mathbf{\omega})$ refers to the CNN feature extractor output based on weight $\mathbf{\omega}$ and the sample $x_i$. This term encourages DNN to learn linear behaviors from samples to features.

Further, the sample-to-label mixup regularizer based on teacher and student models on source domain is defined as
\begin{equation}
    \begin{aligned}
         L^\mathrm{FC}(\mathbf{\omega},\mathbf{\omega}_t)
         = \underset{\lambda\sim\mathrm{Beta}(\alpha,\alpha)}{\mathbb{E}}\ \underset{x_i,x_j\sim\mathbf{D}^\mathrm{src}}{\mathbb{E}} ||z'_\mathrm{src}\left(\mathrm{Mix}_\lambda(x_i, x_j);\mathbf{\omega}\right)
        - \mathrm{Mix}_\lambda({z}''_\mathrm{src}(x_i;\mathbf{\omega}_{t}), z''_\mathrm{src}(x_j;\mathbf{\omega}_{t}))||_2^2\ ,\label{eq:kd}
    \end{aligned}
    \end{equation}
where $z'_\mathrm{src}(x_i;\mathbf{\omega})$ refers to the classifier (Fully-Connected) output of the student model with $x_i$ on $\mathrm{\omega}$ and $z''_\mathrm{src}(x_i;\mathbf{\omega}_t)$ refers to the classifier output of the teacher model with $x_i$ on $\mathrm{\omega}_t$. Both classifiers $z'_\mathrm{src}$ and $z''_\mathrm{src}$ are in the source domain (e.g., with logit outputs in 1,000 dimensions when the model is pre-trained using ImageNet). More specifically, the FC layers in $z'_\mathrm{src}$ and $z''_\mathrm{src}$ are also initialized with the weights of the FC layer in the pre-trained model $\omega_\mathrm{src}$, while $z'_\mathrm{src}$ in the student model is updated for every iteration and $z''_\mathrm{src}$ is updated from the student model. 

The vanilla sample-to-label mixup regularizer $L^\mathrm{MXP}(\omega)$ is derived from the standard implementation of mixup strategy~\cite{zhang2018mixup}  based on student model using the target dataset $\mathbf{D}^\mathrm{tgt}$.

\section{Experiments}
We evaluate our method on a wide range of tasks, covering different kinds of datasets, pre-trained models, data scales and model architectures. Through exhaustive experiments, \TheName\ is compared against multiple state-of-the-art fine-tuning algorithms including $\mathrm{L^2}$~\cite{donahue2014decaf}, $\mathrm{L^2}-SP$~\cite{li2018explicit}, DELTA~\cite{li2019delta}, BSS~\cite{chen2019catastrophic}, RIFLE~\cite{li2020rifle} and Co-Tuning~\cite{you2020co}. 

\subsection{Image Classification}
\subsubsection{Datasets and Models}
We conduct experiments on three popular object recognition datasets:  CUB-200-2011~\cite{wah2011caltech}, Stanford Cars~\cite{KrauseStarkDengFei-Fei_3DRR2013} and FGVC-Aircraft~\cite{Maji2013FineGrainedVC}, which are intensively used in state-of-the-art transfer learning literatures~\cite{chen2019catastrophic,li2020rifle,you2020co}. We use ImageNet~\cite{Deng2009ImageNetAL} pre-trained ResNet-50~\cite{he2016deep} as the source model. For each dataset, we create three subsets which respectively sampling 15\%, 30\% and 50\% training examples from the entire training set, in additional to the original dataset (100\%). These sorts of datasets have been paid more attention recent years due to their realistic scenarios and higher resolution ratios in comparison with traditional datasets such as CIFAR-10 and Caltech256.

Besides these standard tasks, we also compare our method with competitive baselines on more task types and architectures. Specifically, we use the Places365~\cite{zhou2017places} pre-trained ResNet-50 to perform fine-tuning on MIT-Indoors-67~\cite{quattoni2009recognizing} which is a scene classification task. We also investigate how these methods behave on more powerful architecture EfficientNet-B4~\cite{tan2019efficientnet} designed by NAS over a large scale dataset Food-101~\cite{bossard14}.

Information about these datasets is summarized in Table~\ref{tab:dataset} in Appendix. 

\subsubsection{Training Details}
We apply standard data augmentation strategies for image pre-processing composed of random flipping and random cropping to $224\times224$ during training. For inference, the test image is center cropped. we do not use post-processing methods such as ten-crop ensemble. We train all models using SGD with the momentum of 0.9, weight decay of 1e-4 and batch size of 48. We train 16,000 iterations for Food-101 considering its large scale and 9,000 iterations for the remaining datasets. The initial learning rate is set to 0.001 for MIT-Indoor-67 due to its high similarity with the pre-trained dataset Places365 and 0.01 for the remaining. The learning rate is divided by 10 after two-thirds of total iterations. Each experiment is repeated five times and we report the average top-1 classification accuracy and standard division.

For hyperparameter search, we use a simple three-fold cross validation on the training set of CUB-200-2011 from $\gamma_\mathrm{FE} \in$ [0.01, 0.1] and $\gamma_\mathrm{FC} \in$ [0.01, 0.1]. The selected best configurations $\gamma_\mathrm{FE}=0.01$ and $\gamma_\mathrm{FC}=0.1$ are used across all datasets.  As for baseline methods, we use the recommended choices of hyper-parameters reported in their papers. 

\subsubsection{Results}
As observed in Tabel~\ref{table:accu1}, our proposed \TheName\ achieves remarkable improvements to vanilla fine-tuning on three standard benchmarks, and gives the best performance among all state-of-the-art methods. As the sampling rate becomes smaller, our method yields more significant benefits, e.g. \TheName\ outperforms vanilla fine-tuning by 7\% on FGVC-Aircraft when only 15\% training samples are used.

Datasets in Table~\ref{table:accu2} bring more challenges for transfer learning algorithms due to their intrinsic characteristics. For MIT-Indoor-67, vanilla fine-tuning with a small learning rate is quite competitive as the pre-trained model is highly adaptable for the target task. While for large-scale dataset Food-101, the benefit from all fine-tuning algorithms becomes less. In these tasks, \TheName\ still delivers decent performance. 

\begin{table*}
\vspace{-3mm}
  \caption{Comparison of top-1 accuracy (\%) with various algorithms on standard transfer learning benchmarks ($^*$ failed to reproduce the reported accuracy in their paper).}
  \label{table:accu1}
  \centering
  \small
  \begin{threeparttable}
  \begin{tabular}{llllll}
    \toprule
    \multirow{2}*{Dataset} & \multirow{2}*{Method} & \multicolumn{4}{c}{Sampling Rates} \\
    \cmidrule{3-6}
     ~ & ~ & 15\% & 30\% & 50\% & 100\%  \\
    \midrule
     \multirow{7}*{CUB-200-2011} & $\mathrm{L^2}$~\cite{donahue2014decaf} & 43.91$\pm$0.31 & 65.48$\pm$0.22 & 74.08$\pm$0.42 & 79.85$\pm$0.40 \\ 
     ~ & $\mathrm{L^2}$-SP~\cite{li2018explicit} & 44.90$\pm$0.64 & 64.21$\pm$0.82 & 73.80$\pm$0.63 & 79.95$\pm$0.36 \\ 
     ~ & DELTA~\cite{li2019delta} & 48.01$\pm$0.64 & 63.35$\pm$0.33 & 72.82$\pm$0.29 & 81.36$\pm$0.11 \\ 
     ~ & BSS~\cite{chen2019catastrophic} & 46.90$\pm$0.38 & 65.40$\pm$0.22 & 73.10$\pm$0.41 & 80.14$\pm$0.47 \\ 
     ~ & RIFLE~\cite{li2020rifle} & 41.99$\pm$0.82 & 63.43$\pm$0.80 & 73.68$\pm$0.40 & 81.54$\pm$0.10 \\ 
     ~ & Co-Tuning~\cite{you2020co} & 49.26$\pm$0.24 & 66.72$\pm$0.18 & 75.12$\pm$0.18 & 81.88$\pm$0.16 \\ 
     ~ & \TheName\ & \textbf{50.34}$\pm$0.69 & \textbf{69.47$\pm$0.23} & \textbf{76.75$\pm$0.20} & \textbf{82.49$\pm$0.21} \\ 
    \midrule
     \multirow{7}*{Stanford-Cars} & $\mathrm{L^2}$~\cite{donahue2014decaf} & 44.35$\pm$0.63 & 68.96$\pm$0.39 & 82.39$\pm$0.15 & 89.46$\pm$0.19 \\ 
     ~ & $\mathrm{L^2}$-SP~\cite{li2018explicit} & 41.12$\pm$0.24 & 66.96$\pm$0.43 & 80.54$\pm$0.03 & 88.58$\pm$0.21 \\ 
     ~ & DELTA~\cite{li2019delta} & 43.71$\pm$0.76 & 68.39$\pm$0.65 & 81.98$\pm$0.45 & 89.61$\pm$0.26 \\ 
     ~ & BSS~\cite{chen2019catastrophic} & 47.11$\pm$0.74 & 71.79$\pm$0.13 & 83.14$\pm$0.32 & 89.66$\pm$0.12 \\ 
     ~ & RIFLE~\cite{li2020rifle} & 45.45$\pm$0.57 & 71.25$\pm$0.26 & 82.92$\pm$0.38 & 90.08$\pm$0.17 \\ 
     ~ & Co-Tuning~\cite{you2020co} & 42.82$\pm$0.42 $^*$ & 68.68$\pm$0.22 & 82.91$\pm$0.03 & 90.03$\pm$0.06 \\ 
     ~ & \TheName\ & \textbf{50.37$\pm$0.39} & \textbf{72.99$\pm$0.24} & \textbf{84.70$\pm$0.24} & \textbf{91.17$\pm$0.14} \\ 
    \midrule
     \multirow{7}*{FGVC-Aircraft} & $\mathrm{L^2}$~\cite{donahue2014decaf} & 44.47$\pm$0.76 & 66.37$\pm$0.56 & 76.16$\pm$0.65 & 82.83$\pm$0.23 \\ 
     ~ & $\mathrm{L^2}$-SP~\cite{li2018explicit} & 43.20$\pm$0.27 & 64.93$\pm$0.64 & 74.69$\pm$1.04 & 82.60$\pm$0.44 \\ 
     ~ & DELTA~\cite{li2019delta} & 45.13$\pm$0.78 & 67.37$\pm$0.44 & 75.76$\pm$0.16 & 84.45$\pm$0.31 \\ 
     ~ & BSS~\cite{chen2019catastrophic} & 45.73$\pm$0.98 & 67.53$\pm$1.18 & 75.98$\pm$0.55 & \textbf{84.95$\pm$0.05} \\ 
     ~ & RIFLE~\cite{li2020rifle} & 45.87$\pm$0.62 & 65.83$\pm$0.24 & 75.52$\pm$0.44 & 84.37$\pm$0.32 \\ 
     ~ & Co-Tuning~\cite{you2020co} & 45.56$\pm$0.53 & 64.17$\pm$0.40 & 75.45$\pm$0.17 & \textbf{84.84$\pm$0.22} \\ 
     ~ & \TheName\ & \textbf{51.47$\pm$0.62} & \textbf{69.23$\pm$0.56} & \textbf{77.90$\pm$0.57} & \textbf{84.89$\pm$0.23} \\ 
    \bottomrule
  \end{tabular}
  \end{threeparttable}
\vspace{-3mm}
\end{table*}

\begin{table*}
  \caption{Comparison of top-1 accuracy (\%) with different transfer learning algorithms on more task types and architectures. }
  \label{table:accu2}
  \centering
  \small
  \begin{tabular}{lllll}
    \toprule
    \multirow{2}*{Dataset} & \multirow{2}*{Method} & \multicolumn{3}{c}{Sampling Rates} \\
    \cmidrule{3-5}
     ~ & ~ & 30\% & 50\% & 100\%  \\
    \midrule
    \multirow{5}*{MIT-Indoor-67} & $\mathrm{L^2}$~\cite{donahue2014decaf} & 77.94$\pm$0.39 & 80.87$\pm$0.30 & 82.92$\pm$0.32 \\ 
     ~ & DELTA~\cite{li2019delta} & 79.60$\pm$1.00 & 81.85$\pm$0.37 & 83.31$\pm$0.41 \\ 
     ~ & BSS~\cite{chen2019catastrophic} & 76.20$\pm$0.28 & 79.02$\pm$0.56 & 81.64$\pm$0.26 \\ 
     ~ & RIFLE~\cite{li2020rifle} & 76.20$\pm$0.61 & 78.53$\pm$0.36 & 81.58$\pm$0.07  \\ 
     ~ & SMILE & \textbf{81.51$\pm$0.44} & \textbf{83.11$\pm$0.20} & \textbf{84.99$\pm$0.08} \\ 
    \midrule
     
    \multirow{5}*{Food-101} & $\mathrm{L^2}$~\cite{donahue2014decaf} & 80.25$\pm$0.28 & 83.43$\pm$0.15 & 86.77$\pm$0.03 \\ 
     ~ & DELTA~\cite{li2019delta} & 80.58$\pm$0.08 & 83.27$\pm$0.06 & 86.75$\pm$0.02 \\ 
     ~ & BSS~\cite{chen2019catastrophic} & 80.29$\pm$0.11 & 83.30$\pm$0.09 & 86.84$\pm$0.09 \\ 
     ~ & RIFLE~\cite{li2020rifle} & 81.13$\pm$0.04 & 83.82$\pm$0.02 & 86.89$\pm$0.11 \\ 
     ~ & SMILE & \textbf{81.81$\pm$0.14} & \textbf{84.41$\pm$0.16} & \textbf{87.36$\pm$0.03} \\ 
    \bottomrule
  \end{tabular}
\vspace{-3mm}
\end{table*}

\subsection{Natural Language Processing}
We also evaluate \TheName\ on the text classification task using powerful transformer-based architecture, showing that our method can be applied to NLP tasks. 
We use SST-5, which is the Stanford Sentiment Treebank with five categories, as the benchmark. The model is base BERT~\cite{devlin2018bert} with 12 transformer blocks and 12 attention heads. 
We fine-tune the pre-trained BERT model with the batch size to 24 for 3 epochs, using Adam optimizer with lr = 2e-5. 
As shown in Table~\ref{tab:nlp} (in Appendix), we find that both mixup and \TheName\ outperform standard fine-tuning and \TheName\ achieves more improvements. Regularizers L2-SP and BSS without mixup are not superior to standard fine-tuning in this task. 

\subsection{Ablation Study}
We here present an ablation study to exhibit the individual contribution corresponding to each component in the \TheName\ framework. Specifically, we evaluate the performances of independently using sample-to-feature mixup (denoted by M-FE) and sample-to-label mixup on the FC layer (denoted by M-FC) corresponding to the source domain. As observed in Table~\ref{table:ablation}, we find both of them make non-trivial contributions, while the influence of M-FE is dominant.

As for Mean Teacher, we test two extreme strategies without moving average. We denote \emph{w/o S-Teacher} as using the latest target model without introducing the pre-trained model for knowledge distillation and \emph{w/o T-Teacher} as using fixed pre-trained model as the teacher respectively. Results in Table~\ref{table:ablation} show that, both of them incur accuracy drops and it's reasonable to balance the generalization and adaptation by a moving average manner. 

We also evaluate fine-tuning with the standard mixup, denoted by Degenerated-\TheName\ (D-\TheName), which only enforces linear behaviors on the output layer of the target model. D-\TheName\ is equivalent to removing both Feature Mix, Label Mix, and consequently Mean Teacher from \TheName\. But the benefit of D-\TheName\ compared to vanilla fine-tuning is marginal, especially for the 30\% setting. We will diagnose more about this in the next section. 

Moreover, we compare \TheName\ with a similar and more straightforward approach: the combination of D-\TheName\ and knowledge distillation, denoted by D-\TheName+KD. We employ DELTA~\cite{li2019delta} as the knowledge distillation method. Different from \TheName, D-\TheName+KD tends to align the features with respect to mixed inputs from the source model, rather than imitating the proportionally mixed features with respect to original inputs. As shown in Table~\ref{table:ablation}, such direct combination can hardly benefit mixup without explicitly enforcing linear behaviors through deep features. Similarly, directly combining mixup with RIFLE~\cite{li2020rifle} can neither achieve satisfied results. 

\begin{table}
  \caption{Ablation Study on the CUB-200-2011 dataset. FT refers to standard fine-tuning with $\mathrm{L^2}$ regularization.}
  \label{table:ablation}
  \centering
  \small
  \begin{tabular}{lll}
    \toprule
    \multirow{2}*{Method} & \multicolumn{2}{c}{Sampling Rates} \\
    \cmidrule{2-3}
     ~ & 30\% & 100\%  \\
    \midrule
    FT & 65.48$\pm$0.22 & 79.85$\pm$0.40 \\
    FT w/ M-FC & 66.49$\pm$0.59 & 81.61$\pm$0.08 \\
    FT w/ M-FE & 69.29$\pm$0.15 &  82.08$\pm$0.13 \\
    \TheName\ & 69.47$\pm$0.23 & 82.49$\pm$0.21 \\
    \midrule
    \TheName\ w/o S-Teacher & 69.05$\pm$0.37 & 82.35$\pm$0.12 \\
    \TheName\ w/o T-Teacher & 68.66$\pm$0.15 & 82.28$\pm$0.12 \\
    \midrule
    D-\TheName\ & 65.84$\pm$0.15 & 80.77$\pm$0.18 \\
    D-\TheName\ + KD & 66.07$\pm$0.56 & 81.11$\pm$0.06 \\
    D-\TheName\ + RIFLE & 66.46$\pm$0.94 & 81.40$\pm$0.19 \\
    \bottomrule
  \end{tabular}
\end{table}

\section{Discussions}
In order to investigate the inherent mechanism that endows advantages to \TheName, we diagnose why naive mixup fails in situations of transfer learning. In the following subsections, we first formulate a quantitative indicator of the linear behavior which is derived straightly from the training objective of mixup. Then we figure out the existence of \emph{interpolation over-fitting} in fine-tuning with naive mixup and further indicate the necessity of encouraging linear behaviors through deep features. We also explain why the same method imposed on the output layer of the source model is helpful. 
\subsection{Quantifying the Interpolation Loss}
Derived from standard mixup~\cite{zhang2018mixup}, we introduce a generalized form of interpolation loss ($\mathrm{IL}$) w.r.t a function $f$ employing its own outputs as labels, eliminating the influence of the faithfulness of the approximation, i.e. how the learned function fits the underlying ground truth $f^*$, as follows:
\begin{equation}
\label{eq:linearity}
\small
\begin{split}
\mathrm{IL}(f) = & \mathbb{E}_{x,y \sim D} \mathbb{E}_{x\prime,y\prime \sim D} \mathbb{E}_{\delta_1,\delta_2 \sim P_{\delta}} \mathbb{E}_{\lambda \sim P_{\lambda}} 
 D^{it}_{\lambda}(f(\mathrm{Mix}_{ \lambda * \delta_1 + (1- \lambda)* \delta_2 }(x, x\prime),  
 f(\mathrm{Mix}_{\delta_1}(x,x\prime), f(\mathrm{Mix}_{\delta_2}(x,x\prime)
)),
\end{split}
\end{equation}
where $D^{it}_{\lambda}$ refers to the normalized distance between the output w.r.t the interpolated inputs and the proportionally mixed outputs, defined as
\begin{equation}
\label{eq:dist}
\small
\begin{split}
D^{it}_{\lambda}(y_{it}, y_1, y_2) = \frac{D(y_{it}, \lambda*y_1+(1-\lambda)*y_2)}{D(y_1, y_2)}.
\end{split}
\end{equation}

The denominator is used to ensure the comparability between different pairs of $y_1$ and $y_2$ by making the loss independent the scale of the distance between them. In condition of using the ground truth label, the training objective of mixup can be regarded as a special case of minimizing Eq~\ref{eq:linearity} where $\delta_1 = 1, \delta_2 = 0$ and $P_{\lambda} = Beta(\alpha, \alpha)$. 
We would clarify that the linear behavior here is only considered in the interpolation space in-between samples. Related and different topics include the piece-wise linearity~\cite{arora2016understanding} of DNN defined in the entire function space and local linear approximation~\cite{ribeiro2016should} around a specific input point.  

\subsection{Does sample-to-feature mixup help?}

\subsubsection{Fine-tuning overfits to the vanilla mixup}
We first reveal the phenomenon of interpolation over-fitting caused by fine-tuning with vanilla mixup. We leverage Eq~\ref{eq:linearity} to evaluate the label interpolation loss ($\mathrm{IL}$ on the output layer) for different models. Specifically, we use a uniform distribution $U(0.5, 1)$ as $P_{\delta}$ to sample each pair of $\delta_1$ and $\delta_2$. Enforcing both $\delta_1$ and $\delta_2$ larger than 0.5 suggests that we only require linear behaviors in the interpolation region where the same sample ($x$ in Eq~\ref{eq:linearity}) dominates the output w.r.t the mixed input. 
We sample $\lambda$ from the uniform distribution $U(0, 1)$ for interpolation between $\mathrm{Mix}_{\delta_1}(x,x\prime)$ and $\mathrm{Mix}_{\delta_2}(x,x\prime)$ in Eq~\ref{eq:linearity}.

According to Eq~\ref{eq:linearity}, we calculate the $\mathrm{IL}$ w.r.t the output layer for standard fine-tuning and fine-tuning with mixup over both the training and testing set. 30\% of training examples of CUB-200-2011 are used for training. As shown in Table~\ref{tab:loi_output}, although explicitly pursuing linear behaviors during training, mixup causes severe over-fitting that $\mathrm{IL}$ is much higher on the testing set. While the linear behaviors of standard fine-tuning and our proposed \TheName\ generalize well on the testing set.

\begin{table}[h]
\vspace{-3mm}
  \caption{Label interpolation loss for different fine-tuning methods over the training (sampling the CUB-200-2011 training set by 30\%) and testing dataset. Lower is better.}
  \label{tab:loi_output}
  \centering
  \small
  \begin{tabular}{lll}
    \toprule
    \multirow{2}*{Method} & \multicolumn{2}{c}{Dataset} \\
    \cmidrule{2-3}
     ~ & Training & Testing  \\
    \midrule
    Fine-tune & 1.91 & 1.86 \\
    Fine-tune + mixup & 1.65 & 2.00 \\
    \TheName\ & 1.82 & 1.75 \\
    \bottomrule
  \end{tabular}
\vspace{-3mm}
\end{table}

\subsubsection{Sample-to-feature mixup ensures generalizable linear behaviors in both feature and label spaces}
Now we turn to feature interpolation loss, which may faithfully reflect the linear behaviors of CNN feature extractors. In our experiments, we chose the last hidden layer in ResNet-50 with 2048 activation maps composed. We follow the same settings in 5.2.1 and quantify the feature interpolation loss. Several arguments can be deduced from results in Table~\ref{tab:il_feature}. 

\emph{A. Fine-tuning with mixup shows non-generalizable label interpolation behaviors because it fails to obtain strong linear behaviors on features.} As shown in Table~\ref{tab:il_feature}, feature interpolation losses on both training and testing dataset are very high for mixup. Weak linear behaviors on deep features will certainly lead to high label interpolation loss on unseen examples. 

\emph{B. Linear behaviors on features relates more with generalization than memorization.} When we involve additional training examples, the feature interpolation loss for the original training set does not rise. This is a non-trivial phenomenon because additional samples compete with original samples on interpolation memorization given the same training budgets. Further, utilizing more training data, which is undoubtedly the most reliable manner to promote generalization on unseen data, leads to stronger linear behaviors on features. 

\emph{C. Linear behaviors on features is not prone to over-fitting.} In all experiments, feature interpolation losses between the training and testing set have similar values. More importantly, explicitly minimizing the feature interpolation loss in the training set still brings similar benefits to the testing set. 

These arguments solidify our motivation of sample-to-feature mixup.

\begin{table}[h]
\vspace{-3mm}
  \caption{Feature interpolation loss for different fine-tuning methods over the training (sampling the CUB-200-2011 training set by 30\%) and testing dataset. Lower is better. Add. Data refers to involving the remaining 70\% training examples for fine-tuning. While the interpolation loss for the training set is still calculated on the original 30\%.}
  \label{tab:il_feature}
  \centering
  \small
  \begin{tabular}{lll}
    \toprule
    \multirow{2}*{Method} & \multicolumn{2}{c}{Dataset} \\
    \cmidrule{2-3}
     ~ & Training & Testing  \\
    \midrule
    Feature Extractor & 1.93 & 1.92  \\
    Fine-tune & 1.58 & 1.66 \\
    Fine-tune + Add. Data & 1.58 & 1.63 \\
    Fine-tune + mixup & 1.98 & 2.02 \\
    \TheName\ & 1.48 & 1.53 \\
    \bottomrule
  \end{tabular}
\vspace{-3mm}
\end{table}

\subsection{Does sample-to-label mixup in the source domain help fine-tuning in the target domain?}
Assuming that the label space of the source task is at least partial related with that of the target task, the FC layer with a considerable number of parameters contains useful information for the target task. This has been observed and exploited by relating the label space between these two tasks in recent studies~\cite{you2020co}.

Our work aims to encourage linear behaviors on the feature space by leveraging the feature-to-label classifier in the source model. Since there is no real ground truth for feature interpolation, supervisions from the label space, which has been well trained over the large-scale source dataset, provides additional signals which is complementary to directly reducing feature interpolation loss through a mean teacher. 
\section{Conclusion}

In this work, we introduce \TheName---Self-distilled Mixup strategies for Efficient Transfer Learning. Beyond a direct combination of fine-tuning and mixup, \TheName\ incorporates a mean teacher-student training framework with triplet regularization, encouraging linear behaviors and cross-domain generalizability. We conduct extensive experiments using a wide spectrum of target datasets. Results show that \TheName\ can significantly promote the effectiveness of fine-tuning and outperform various competitive fine-tuning algorithms. Ablation studies and empirical discussions further backup our design intuition and purposes.

{
\bibliography{references}
\bibliographystyle{unsrt}
}
\appendix

\clearpage
\section{Descriptions about the benchmarks.}
The descriptions about the benchmarks used in image classification tasks are summarized in Table~\ref{tab:dataset}.

\begin{table}[h]
  \caption{Characteristics of the target tasks.}
  \label{tab:dataset}
  \centering
  \begin{tabular}{llllll}
    \toprule
    target dataset & task category & source task & architecture & \# training & \# classes \\
    \midrule
    CUB-200-2011 & object recognition & ImageNet & ResNet-50 & 5,994 & 200 \\
    Stanford-Cars & object recognition & ImageNet & ResNet-50 & 8,144 & 196 \\
    FGVC-Aircraft & object recognition & ImageNet & ResNet-50 & 6,677 & 100 \\
    MIT-Indoor-67 & scene classification & Places365 & ResNet-50 & 5,356 & 76 \\
    Food-101 & object recognition & ImageNet & EfficientNet-B4 & 75,000 & 101 \\
    \bottomrule
  \end{tabular}
\end{table}

\section{Experiments on SST-5}
The NLP experiment is performed on the fine-grained sentiment classification task SST-5. We use the standard training configurations and report the accuracies on classic architectures LSTM and CNN according to previous work~\cite{munikar2019fine}. Then we reproduce the experiment using BERT$_{base}$ and evaluate different transfer learning algorithms. Results are shown in Table~\ref{tab:nlp}.

\begin{table}[h]
  \caption{Experimental results on NLP task SST-5. }
  \label{tab:nlp}
  \centering
  \small
  \begin{tabular}{ll}
    \toprule
    Method & Accuracy \\
    \midrule
    LSTM~\cite{tai2015improved} & 46.4 \\
    CNN~\cite{kim2014convolutional} & 48.0 \\
    BERT$_{base}$~\cite{devlin2018bert} & 53.2 \\
    BERT$_{base}$+Mixup~\cite{zhang2018mixup} & 53.7 \\
    BERT$_{base}$+$\mathrm{L^2}$-SP~\cite{li2018explicit} & 53.2 \\
    BERT$_{base}$+BSS~\cite{chen2019catastrophic} & 53.4 \\    
    BERT$_{base}$+SMILE & \textbf{54.6} \\    
    \bottomrule
  \end{tabular}
\end{table}

\section{Feature Interpolation Visualizations}
Here we present some visualization plots to observe the interpolation behaviors of different fine-tuning methods on the feature space. 

\begin{figure}
\centering
\includegraphics[width=0.9\textwidth]{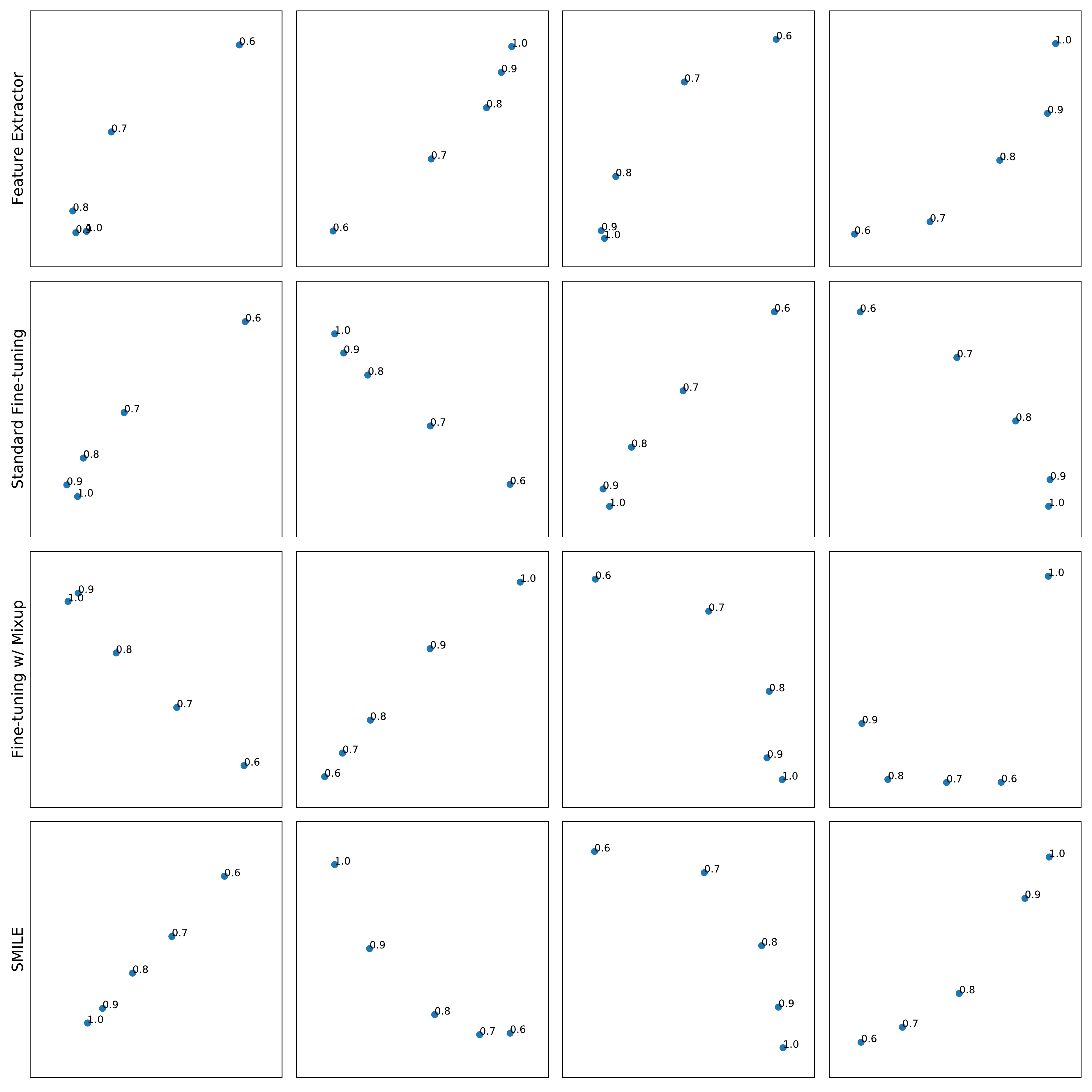} 
\caption{Visualizations of feature interpolation behaviors for different fine-tuning methods. We extract the representations from the last hidden layer which are  2048 dimensional feature vectors and then project them into the 2-D space using PCA. Each column corresponds to the projected deep features generated by putting forward the interpolation of a random pair of images. The number marked next to the point refers to the interpolation coefficient $\lambda$. }
\label{fig:interpolation} 
\end{figure}

\textbf{Methods.} To obtain the interpolation points, we first randomly select a pair of images and then generate five mixed inputs with the interpolation coefficients of [0.6, 0.7, 0.8, 0.9, 1] respectively. Forward computation is performed given these mixed inputs and then, their corresponding deep features are extracted and projected to the 2-D space using PCA. Results of four random pairs are illustrated in Figure~\ref{fig:interpolation}. 

\textbf{Observations.} 
As shown in Fig~\ref{fig:interpolation}, the pre-trained model without fine-tuning on the target dataset (the top row) sometimes exhibits rather strong linear behaviors (columns 2 and 4). Sometimes the outputs tend to saturate when $\lambda$ is near 1 (columns 1 and 3), i.e. the outputs are very close to each other. Standard fine-tuning (the 2rd row) seems to be capable of alleviating the saturation problem but do not bring remarkable changes in linear behaviors. While fine-tuning with mixup (the 3rd row) sometimes dramatically destroy the existing linear structures among the interpolated points (column 4). \TheName\ (the 4th row) shows significantly stronger linear behaviors compared against the feature extractor and other fine-tuning strategies.

\end{document}